\documentclass[runningheads]{llncs}

\usepackage{eccv}

\usepackage{eccvabbrv}

\usepackage{graphicx}
\usepackage{booktabs}
\usepackage[table]{xcolor}
\usepackage[accsupp]{axessibility}  
\usepackage{multirow}
\usepackage{algorithm}
\usepackage{algpseudocode}

\usepackage{hyperref}

\usepackage{orcidlink}

\begin{document}

\title{StreamingVLA: Streaming Vision-Language-Action \\ Model with Action Flow Matching and Adaptive Early Observation} 

\titlerunning{StreamingVLA}

\author{Yiran Shi\inst{1} \and Dongqi Guo\inst{1} \and Tianchen Zhao \inst{1} \and FengGao \inst{1} \and Liangzhi Shi \inst{1} \and Chao Yu \inst{1} \and ZhiJian Mo \inst{2} \and Qihua Xiao \inst{2} \and XiaoShuai Peng \inst{2} \and Qingmin Liao \inst{1} \and Yu Wang \inst{1}}

\authorrunning{Y.~Shi et al.}

\institute{Tsinghua University \and
Lenovo Group Ltd.}

\maketitle
\begin{abstract}
Vision-language-action (VLA) models have demonstrated exceptional performance in natural language-driven perception and control. However, the high computational cost of VLA models poses significant efficiency challenges, particularly for resource-constrained edge platforms in real-world deployments. 
However, since different stages of VLA (observation, action generation and execution) must proceed sequentially, and wait for the completion of the preceding stage, the system suffers from frequent halting and high latency.
To address this, We conduct a systematic analysis to identify the challenges for fast and fluent generation, and propose enabling VLAs with the ability to asynchronously parallelize across VLA stages in a \textbf{"streaming"} manner. First, we eliminate the reliance on action chunking and adopt action flow matching, which learns the trajectory of action flows rather than denoising chunk-wise actions. It overlaps the latency of action generation and execution. Second, we design an action saliency-aware adaptive observation mechanism, thereby overlapping the latency of execution and observation. Without sacrificing performance, StreamingVLA achieves substantial speedup and improves the fluency of execution. It achieves a 2.4 $\times$ latency speedup and reduces execution halting by 6.5 $\times$. 
\end{abstract}    
\section{Introduction}
\label{sec:intro}

\begin{figure}[th]
    \centering
    \includegraphics[width=1.0\linewidth]{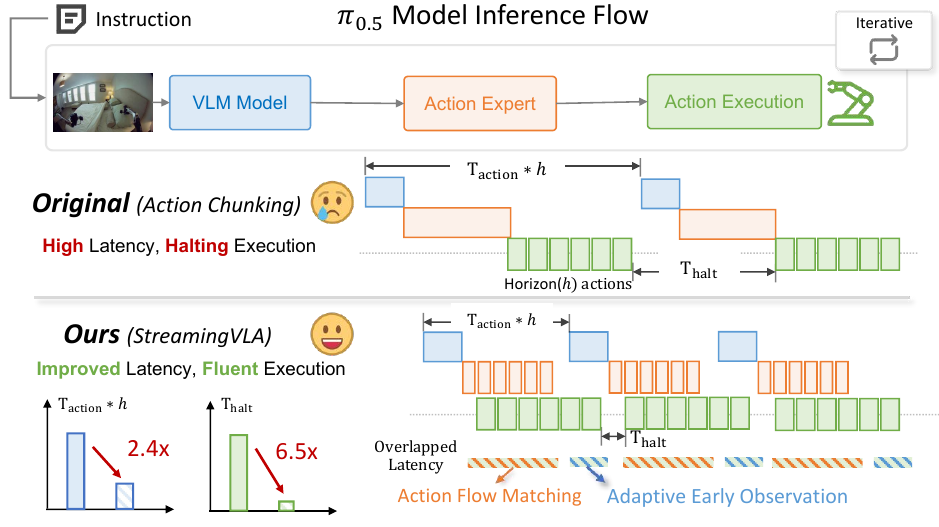}
    \caption{\textbf{StreamingVLA} explores streaming execution of VLA models by enabling different stage to be runned in a asynchronous manner. By overlapping the latency for stages, without sacrificing performance, it achieves \textbf{2.4$\times$} end-to-end speedup, and \textbf{6.5 $\times$} execution halting time reduction, enabling fast and fluent execution. }
    \label{fig:teaser}
\end{figure}

Embodied systems require strong generalization capabilities to operate effectively across diverse real-world scenarios. Inspired by the recent success of foundation models, such as large language models (LLMs)~\cite{LLM_effect_on_robotics,LLM+robotics} and vision-language models (VLMs)~\cite{VLM_survey}, that can perform a wide range of multimodal tasks, \textbf{vision-language-action (VLA)}~\cite{survey_on_VLAs_,openvla,tracevla,navila,gr-2,3d-vla,o2024open} models have emerged as a promising direction toward building foundation models for embodied agents. Despite their impressive performance, the computational cost of large-scale VLA models poses significant efficiency challenges~\cite{resource_challenges_of_VLAs,OpenHelix}, especially on edge platforms. With the adoption of the action chunking technique~\cite{action_chunking—ACT, one_act_play, action_chunking_as_policy_compression}, which generates chunks of actions in parallel, the latency challenge is alleviated. However, there is still room for further improvement~\cite{VLA_cache,VLA_accelerating_chunkstill}.

Aside from the overall high latency challenge, current VLA execution also suffers from \textbf{inconsecutive and stuttered execution}~\cite{stuttered_spped,VLA_challenges}. Taking the $\pi_{0.5}$ model as an example, it integrates a pre-trained vision-language model (VLM) for observation processing and a diffusion-based action expert for action generation. As illustrated in the timeline diagram in \cref{fig:teaser}, the multi-stage model inference and action execution proceed in a \textbf{synchronous} manner, where each stage must wait for the other to complete. Consequently, a significant waiting period where the system stalls, denoted as $T_{\text{halt}}$, occurs between successive executions, leading to noticeable halting.

A line of research~\cite{training-time-RTC,vlash} explores parallelizing execution with VLM-based observation processing to address the aforementioned issue. To further improve performance and gain a clearer understanding of the latency and stalling challenges in VLA execution, as well as the potential of \textbf{asynchronous and parallelized ``streaming'' execution across stages}, we conduct a systematic runtime analysis of multi-stage VLAs (\cref{sec:method-runtime-analysis}).
Based on this analysis, we first explore a new dimension of parallelized execution across action generation and execution. We introduce \textbf{state-based modeling of action flow matching} (\cref{sec:method-action-flow-matching}), which reformulates action modeling: instead of learning an absolute value mapped from noise, the prediction updates a state conditioned on prior actions, better reflecting the action's physical meaning. This formulation removes the reliance on action chunking and enables each action to be executed immediately upon generation.
Secondly, to further overlap action execution with the next iteration’s observation processing, we identify action saliency, which describes the extent to which an action influences subsequent observations. We find that action saliency varies significantly across actions, and existing techniques that treat each action uniformly may yield suboptimal performance preservation. Accordingly, we design an \textbf{action-saliency-aware adaptive early observation scheme} (\cref{sec:method-early-observation}). We summarize our contribution as follows:


\begin{itemize}
    \item We perform systematic runtime analysis of the VLA model inference and execution, and conclude optimization targets for fast and fluent execution. 
    \item We present the StreamingVLA framework, which enables asynchronous and parallelized processing of multiple stages of VLA system. 
    \item Without sacrificing performance, StreamingVLA achieves 2.4$\times$ speedup, and reduce the halting time by 6.5$\times$ to the minimum, enabling fast and fluent execution.
\end{itemize}


\section{Related work}
\label{rw}

\noindent \textbf{Vision-Language-Action Models: }
Vision-language-action (VLA) models emerge as a promising direction toward building foundation models for embodied action generation. Early advances, such as RT-1 \cite{RT-1}, introduced a model capable of fundamental visual-semantic comprehension for robotic control. RT-2 \cite{RT-2} further extended this paradigm by incorporating vision-language modeling component, enhancing reasoning and generalization. $\pi_0$~\cite{black2024pi0visionlanguageactionflowmodel, pi0.5} model series integrated a pre-trained VLM with a diffusion-based policy, for environment understanding and action generation. It achieves remarkable performance, and attracted research interest. In this work, we apply our techniques to the $\pi_{0.5}$ model to validate its effectiveness.

\noindent \textbf{Model Compression for VLA: } A line of existing research focuses on improving the efficiency of VLAs by adapting model compression techniques. Prior literature~\cite{RLRC, MoLeVLA} applies pruning and quantization to reduce the cost of VLM model, while other studies~\cite{EfficientVLA, FAST, FitPrune, async_vla} focus on token compression or part correction to reduce data to be processed.
Although substantially reduce the latency of each stage for VLA, relying solely on them faces challenges in minimizing action execution haltings (as discussed in \cref{sec:method-runtime-analysis}). We adopt a different perspective, which overlaps the latency for different stages. It could work in parallel with existing techniques and minimize halting to achieve fluent execution.

\noindent \textbf{Parallelized Execution for VLA: }
Action chunking~\cite{action_chunk,action_chunk_2} has been widely adopted, which improves execution efficiency by generating multiple actions in parallel for each observation. RTC~\cite{RTC,training-time-RTC}, VLASH~\cite{vlash} and SmolVLA~\cite{smolvla} introduce early observation mechanism through finetuning or test-time adjustment that performs VLM midway through ongoing action execution, which enables overlapping action execution and model inference. To enhance the early observation with the ability to distinguish diverse action saliency, we introduce an adaptive observation scheme (\cref{sec:method-early-observation}). Beyond early observation, StreamingVLA provides an extensive runtime analysis (\cref{sec:method-runtime-analysis}) and extends parallelized execution to a new dimension by enabling stage-wise parallelism between action generation and execution (\cref{sec:method-action-flow-matching}) with state-based action flow matching inspired by SFP~\cite{streaming_flow_policy}.

\begin{figure*}[t]
    \centering
    \includegraphics[width=1.0\linewidth]{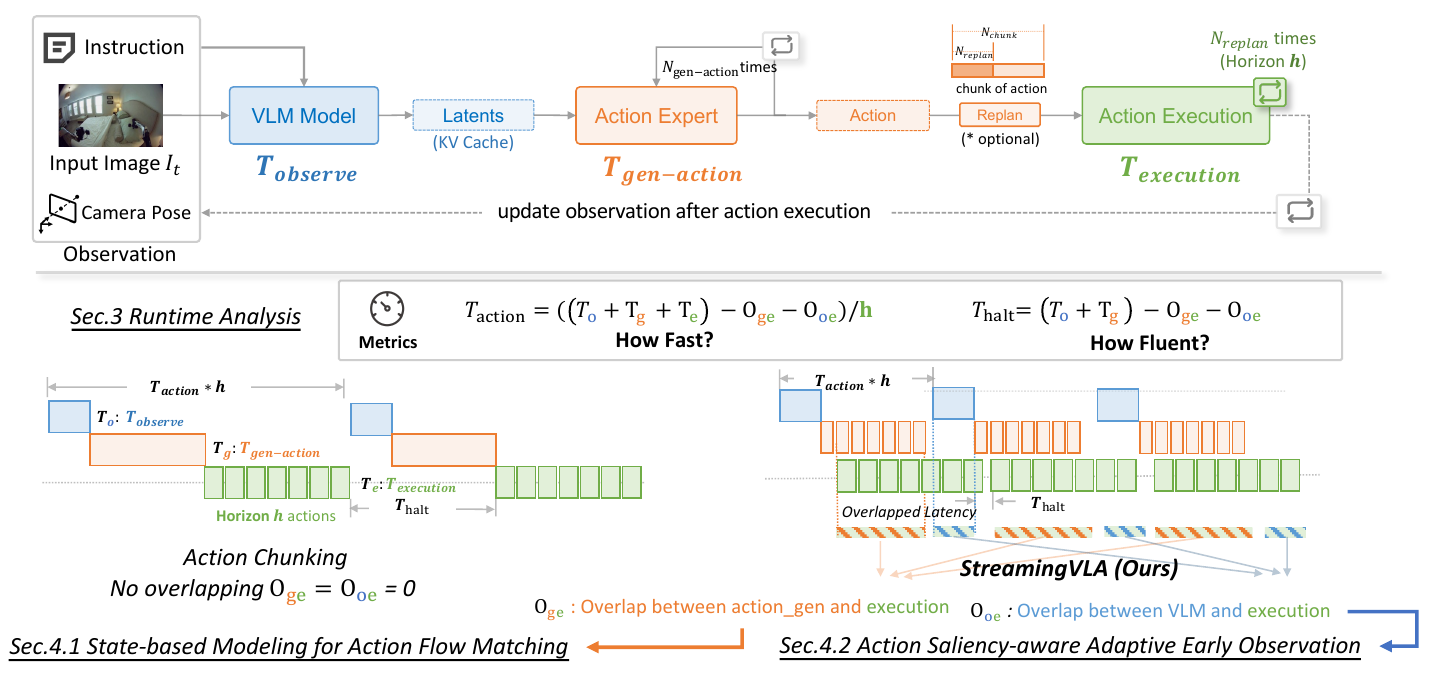}
    \caption{\textbf{The overall methodology framework for StreamingVLA:} We conduct a systematic timeline analysis and conclude the optimization target for fast and fluent execution. We present two key techniques: action flow matching and adaptive early observation, to overlap the latency of action execution with of action generation and VLM observation, respectively.}
    \label{fig:method}
\end{figure*}

\section{Preliminary Analysis of VLA Runtime}
\label{sec:method-runtime-analysis}

To gain a deeper understanding of the issue of VLA's inconsecutive and slow execution, and the potential of asynchronous and parallelized execution across stages. We conduct a systematic \textbf{runtime analysis}. 
As shown in the upper part of \cref{fig:method}, we illustrate the overall workflow of the VLA model’s inference and action execution, along with the corresponding timeline diagrams. The workflow represents VLAs composed of VLM and diffusion expert (such as $\pi_0$ series~\cite{black2024pi0visionlanguageactionflowmodel, pi0.5}, Gr00t~\cite{gr00t}, GR3~\cite{gr3}), we use $\pi_{0.5}$ model as the example, but the analysis is generally applicable. 
The process can be divided into three main stages: First, the ``observation'' stage, where the VLM generates latents (kv-cache) based on the current observation, which includes the camera image, language instruction, and the robot arm’s state. Next, for the ``gen-action'' stage, the KV cache is passed to the action expert, which produces a chunk of actions of size $N_{\text{chunk}}$ through a diffusion process. An optional replan step then selects the first $N_{\text{replan}}$ actions for execution to adjust performance efficiency trade-off. The ``horizon'' $h$ is defined as the number of valid actions for each observation, which equals to $N_{\text{replan}}$. Finally, in ``execution'' stage , these $h$ actions are executed sequentially on the robotic arm. After execution is completed, the observation is updated, and the next iteration begins.

In the aforementioned standard action chunking pipeline, each stage operates synchronously, meaning that every stage must wait for the preceding one to complete. As a result, between consecutive execution stages, there exists a halting gap of $T_{\text{halt}} = T_{\text{observe}} + T_{\text{gen-action}}$, based on actual profiling, the waiting time is substantial, comparable with the whole execution stage's latency. It causes robot arm’s motion to become stuttered and inconsecutive.
The fluency of execution can thus be characterized by $T_{\text{halt}}$, when $T_{\text{halt}}$ approaches zero, the system achieves smooth and fluent execution. Similarly, the latency per action, which reflects the overall execution speed, can be expressed as the sum of the latencies of all stages: $T_{\text{action}} = (T_{\text{observe}}+T_{\text{gen-action}}+T_{\text{execution}})/h$ (abbreviated as $T_o, T_g, T_e$, respectively).

To achieve fast and fluent execution, we aim to minimize both $T_{\text{action}}$ and $T_{\text{halt}}$. Existing model compression techniques~\cite{TinyVLA2024,EfficientVLA} can reduce latency by directly reducing the computational cost of $T_o$, $T_g$. However, it remains challenging to minimize $T_{\text{halt}}$ to near zero value, since it inherently equals $T_o$ + $T_g$.
Instead of directly shortening individual stage latencies, we take a different approach—overlapping the latency of different stages to enable the VLA to operate in an \textbf{asynchronous and streaming} manner. We define $O_{ge}$ and $O_{oe}$ as the overlapped latency between the generation–execution and observation–execution stages, respectively. With these overlaps, the effective latency metrics can be expressed as:
\begin{equation}
\begin{split}
T_{\text{action}} &= ((T_o + T_g + T_e) - (O_{ge} + O_{oe}))/h \\
T_{\text{halt}} &= (T_o + T_g) - (O_{ge} + O_{oe}).
\end{split}
\end{equation}
\label{equ:eff-metrics}
As shown above, the ``streaming'' optimization can work in parallel with model compression techniques that reduce $T_g, T_o$. It further overcomes the limitation of synchronous pipelines, and enables reducing $T_{\text{halt}}$ toward nearly zero.

To achieve $O_{ge}$ and $O_{oe}$, we develop techniques accordingly. As illustrated in the lower part of \cref{fig:method}, we first adapt the streaming action flow matching to the VLA model, enabling it to efficiently generate actions one by one, allowing simultaneous action generation and execution. Secondly, we enhance the early observation mechanism so that the VLM can start processing in advance, before all action executions are completed, thereby achieving concurrent VLM processing and action execution.

\section{Method}
\label{sec:method}

Based on the aforementioned analysis, we propose StreamingVLA, which enables parallelized execution along two dimensions.
First, we introduce a new overlapping dimension between action generation and execution through state-based modeling for action flow matching (\cref{sec:method-action-flow-matching}). This reformulates action modeling from predicting actions as absolute values to predicting actions as updates to a feature-space state that represents the physical system conditioned on prior actions.
Second, to overlap action execution with the next iteration’s VLM observation processing, we introduce an improved early observation scheme (\cref{sec:method-early-observation}) that exploits the diversity of action saliency and adaptively avoids early observation when the action saliency is high.

\begin{figure}
    \centering
    \includegraphics[width=1.0\linewidth]{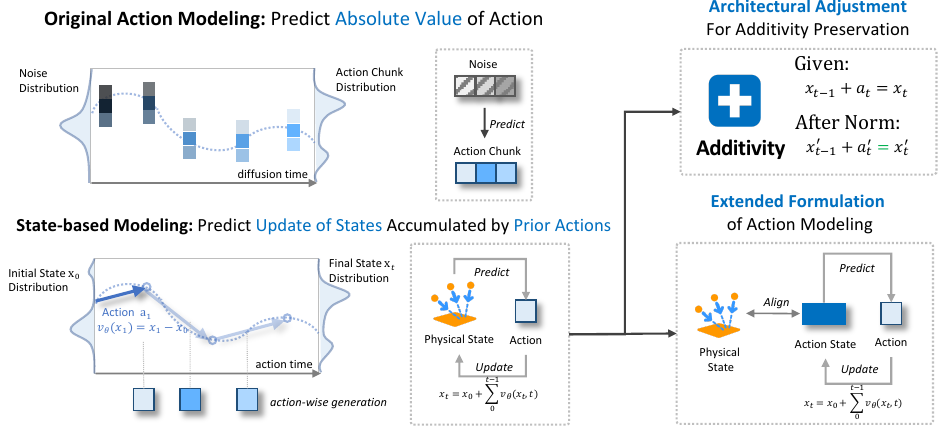}
    \caption{\textbf{The Illustration of state-based modeling of action flow matching:} It reformulates the action modeling from predicting actions as absolute values to predicting actions as updates to a feature-space state accumulated by prior actions. Extended formulation and architectural adjustment is adopted for larger scaled VLA model and mainstream benchmarks.}
    \label{fig:method-afm}
\end{figure}

\subsection{State-based Modeling of Action Flow Matching}
\label{sec:method-action-flow-matching}

In this section, we first present the detailed formulation of action flow matching proposed in Streaming Flow Policy (SFP~\cite{streaming_flow_policy}). We then clarify the nontrivial challenge of adopting such scheme in larger-scaled VLA models and complex tasks. Then, we introduce our corresponding solution of extended formulation, and architectural adjustments.

\subsubsection{Primary Formulation of Action Flow Matching}

In commonly adopted action chunking~\cite{action_chunk} scheme, The action expert of $\pi_{0.5}$ predicts an entire chunk of future actions based on the VLM processed observation $o$. For a noisy action chunk at a given timestep $t\in[0,1]$, denoted as $\mathcal{A}_h^t$, where $h$ implies the horizon (size of the chunk), the model predicts the velocity field $v_{\theta}(\mathcal{A}_h^t,t\vert o)$ along the flow trajectory from the noise distribution $p(\mathcal{A}_h^0)$ to the data distribution $p(\mathcal{A}_h^1)$. The model progressively denoise $\mathcal{A}_h^t$ until the final clean action chunk $\mathcal{A}_h^1$ is obtained. During this process, any intermediate result  $\mathcal{A}_h^t$ contains neither complete nor accurate action information, thus only upon the completion of denoising process can all action tokens within the chunk be determined. 

To enable parallelization between action generation and execution, action-wise generation, rather than chunk-wise generation, is required. A naive approach, equivalent to setting $N_{\text{chunk}}=1$ in action chunking satisfies this requirement, but incurs excessive cost due to iterative denoising for each individual action, motivating the need for a more efficient formulation of action-wise generation.
To address this, we adopt the action flow matching formulation proposed in Streaming Flow Policy (SFP~\cite{streaming_flow_policy}). Instead of denoising chunked actions, this formulation directly models the action trajectory using flow matching. It \textbf{reformulates action modeling from predicting absolute action values from a noise distribution, to predicting updates to the physical state conditioned on prior actions.} This formulation better aligns with the physical semantics of the task and enables efficient one-by-one action generation, facilitating parallelized execution between action generation and execution. Specifically, we train the model to learn the velocity field of the flow constructed by action trajectory. Given an action trajectory $\xi(t)$, we define its conditional action flow as follows:
\begin{equation}
    v_{\xi}(x,t)=\dot{\xi}(t)-k(x-\xi(t)),\ p_{\xi}^0(x) = \mathcal{N}(\xi(0),\sigma_0^2)
\label{eq:velocity field}
\end{equation}
Where $p_{\xi}^0(x)$ denotes the initial distribution of the action states, modeled as a small Gaussian centerd at $\xi(0)$ with variance $\sigma_0^2$. The constant $k$ serves as a stabilizing factor to constrain the behavior of flow when $x$ does not strictly fall on trajectory $\xi(t)$. As demonstrated in SFP, the marginal distribution of $x$ over time is:
\begin{equation}
    p_{\xi}(x\vert t)=\mathcal{N}(\xi(t),\sigma_0^2e^{-2kt})
    \label{eq:distribution}
\end{equation}
Indicating that at any time step $t$, the distribution of $x(t)$ forms a narrow Gaussian centered around $\xi(t)$, with variance exponentially decaying over time. It is also worth noting that the time step $t$ here represents the normalized action index within horizon, instead of denoising timestep. 

For training the action flow matching, from a dataset $p_{\mathcal{D}}$, we sample trajectories and their corresponding raw or processed observations $(\xi,o)\sim p_{\mathcal{D}}$. For each trajectory, we compute the flow velocity field $v_{\xi}(x,t)$ by \ref{eq:velocity field}, the model is trained by minimizing the conditional flow matching loss:
\begin{equation}
\begin{aligned}    
\mathcal{L}(v_{\theta},p_{\mathcal{D}})=& \\
\mathbb{E}_{(o,\xi)\sim p_{\mathcal{D}}}\mathbb{E}&_{t\sim U[0,1]}\mathbb{E}_{x\sim p_{\xi}(x\vert t)}\vert\vert v_{\theta}(x,t\vert o)-v_{\xi}(x,t)\vert\vert_2^2.
\label{eq:flow matching loss}
\end{aligned}
\end{equation}
Once trained, the learned velocity field $v_{\theta}(x,t\vert o)$ can be integrated to obtain the state $x_t$ along the conditional flow:
\begin{equation}
x_t=x_0+\int_0^t v_{\theta}(x(\tau),\tau\vert o)\mathrm{d}\tau.
\end{equation}
The discretization of the integral gives:
\begin{equation}
x_T=x_0+\sum_{t=0}^{T-1}v_{\theta}(x_t,t\vert o) \Delta t,\ T\in\{1,2,\dots,h\}.
\end{equation}
The action we desired can be expressed as $a_t=x_{t+1}-x_t=v_{\theta}(x_t,t\vert o)\Delta t$. Therefore, each action expert forward produces a single velocity estimation $v_{\theta}(x_t,t\vert o)$, which can be immediately converted into one action token and executed, enabling streaming action generation.

Although the aforementioned formulation has been validated in relatively small-scale diffusion policy models and simpler tasks such as PushT~\cite{diffusion_policy}, \textbf{extending it to mainstream benchmarks and large-scale VLA models remains non-trivial} for two main reasons:
\begin{enumerate}
    \item The more complex control schemes in mainstream benchmarks require an extended action modeling formulation for effective adaptation.
    \item The more complex architectures of VLA models must be carefully modified to preserve the essential additivity between actions.
\end{enumerate}

To address these challenges, we propose the following corresponding solutions:

\noindent \textbf{Extended Formulation of State-based Action Modeling:} 
In the original formulation, the state $x_t$ is identical to the robot's physical state. This assumption holds for relatively simple tasks, where direct control logic can be applied. For example, in the PushT task, the policy directly outputs the physical space $x$ and $y$ coordinates for execution. 
However, in more complex mainstream benchmarks, the generated action is passed to an independent controller that indirectly affects the robot's physical state. For instance, in the LIBERO benchmark, the Operational Space Controller (OSC\_POSE) from robosuite converts takes in the policy ouput of end-effector pose deltas and calculates how it affects the whole simulation environment. Due to the nonlinearity introduced by the controller, the action can no longer directly represent the change in the physical state, rendering the original formulation inapplicable.
To address this issue, we extend the formulation by introducing an \textbf{action-space state}, while ensuring its alignment with the corresponding \textbf{physical-space state}.

In practice, generated actions are used to update the action-space state, rather than the physical-space state. To fit such extended formulation and ensure alignment between action and physics space, the training scheme also should be modified accordingly. In the original training procedure, the flow matching process samples different sub-trajectories from the trajectory and performs prediction on these sub-trajectories. 

While the absolute physical state is always accessible in the original formulation, in the extended formulation, the absolute action-space state, which is obtained by accumulating all preceding actions is not directly available during training. Since training only observes sub-trajectories, computing the action state by accumulating actions within merely a sub-trajectory would lead to a mismatch between the action-space state and the corresponding physical-space state, which may cause training failure. Therefore, before training, we precompute the complete action-space states along the trajectory, and input the initial action-space state corresponding to the physical state while training every sub-trajectory to ensure alignment.

In the inference stage, we maintain an action-space state $x_t$ as part of model input, and obtain target velocity field $v_{\theta}(x_t,t)$ after a model forward pass. We take $a_t = v_{\theta}(x_t,t)\Delta t$ as the action output, and update the action-space state using:
\begin{equation}
    x_{t+1} = x_t + v_{\theta}(x_t,t)\Delta t
\end{equation}
thereby consistently preserving the alignment between the action-space state and the physical-space state.

\noindent \textbf{Architectural Adjustment for Additivity Preservation:} 
The key of state-based action modeling scheme is that it maintains a state that is updated through the cumulative of historical actions. The additivity between actions and states is an essential property. In small diffusion policy models, which composed of simple MLPs, naturally satisfy this property. However, in larger-scale VLA models, normalization layers introduced to stabilize training may break this additivity, which can lead to training collapse. 

Specifically, the original normalization is defined as:
\begin{equation}
    a_t' = (a_t - q_{\min})/\text{scale} \times 2 - 1 , \quad \text{scale} = q_{\max} - q_{\min},
\end{equation}
where $q_{\max}$ and $q_{\min}$ are the maximum and minimum values of every dimension, which can be calibrated from the dataset in advance.
When the original actions or states satisfy the additivity property \(x_t + a_t = x_{t+1}\), the normalized variables generally fail to preserve this property, i.e., \(x_t' + a_t' \neq x_{t+1}'\).

To preserve additivity, we introduce two modifications. First, we share the normalization statistics across all action-space states and actions so that the scaling factor remains consistent. Second, we modify the normalization layer by removing the offset term, resulting in:
\begin{equation}
    a_t' = a_t/\text{scale}, \quad \text{scale} = q_{\max} - q_{\min}.
\end{equation}
This simple modification is empirically effective in stabilizing training while preserving the additivity property.

\begin{figure}
    \centering
    \includegraphics[width=0.9\linewidth]{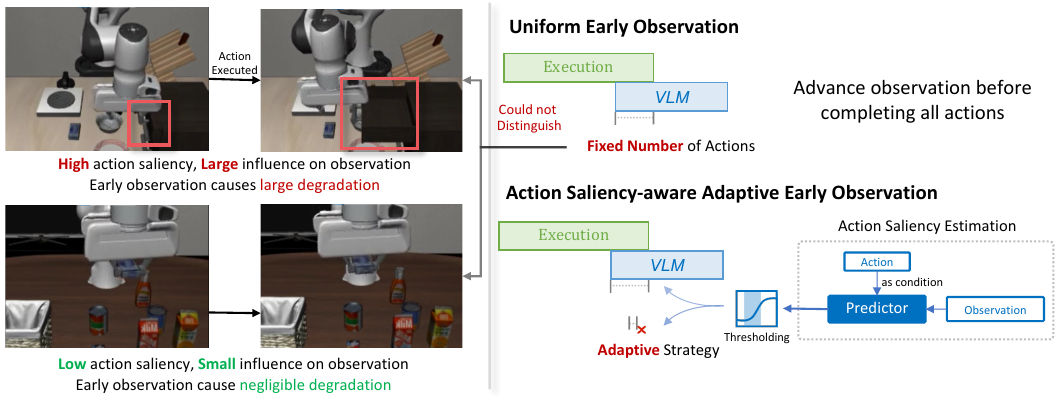}
    \caption{\textbf{The Illustration of action saliency aware adaptive early observation:} We highlight the importance of diverse action saliency, and adopt a lightweight predictor to implement an adaptive early observation scheme. }
    \label{fig:method-eo}
\end{figure}
\subsection{Action Saliency-aware Adaptive Early Observation}
\label{sec:method-early-observation}

The aforementioned action flow matching overlaps most of the action execution with action generation. To further reduce the halting time, we aim to overlap action execution with VLM processing. Early observation, which initiates VLM processing before the completion of the entire action sequence, is well suited for this purpose. Prior work such as RTC~\cite{RTC} and SmolVLA~\cite{smolvla} achieves speedup by advancing observation processing. VLASH~\cite{vlash} further finetunes the model to mitigate the mismatch introduced by early observation. Despite the substantial speedups achieved, noticeable performance degradation still remains. In this work, we aim to further improve performance preservation under early observation.

Through the analysis of failure cases caused by improper early observation, we observe that "action saliency", which reflects the extent to which an action influences the resulting observation, varies significantly across actions. As illustrated in \cref{fig:method-eo}, the upper action completes pulling out the shelf, resulting in a significant environmental change. Performing early observation before this critical change leads to severe errors. In contrast, for most actions (such as the lower example), little observable environmental change occurs, and advance observation before executing these actions will not cause noticeable degradation. Consequently, a uniform early observation strategy that always skips the last few actions in a chunk may discard highly salient actions, leading to notable performance degradation. To address this issue, we propose an action-saliency-adaptive early observation scheme that avoids skipping actions with high saliency.

The key to accurate adaptive early observation is to properly measure action saliency in order to identify important actions. A straightforward approach is to use the norm of the predicted action tokens, as their physical interpretation suggests a positive correlation between token magnitude and movement in physical space. However, our empirical results show that although this strategy improves performance compared with naive uniform early observation, it still leads to noticeable degradation (See \cref{tab:mainres} for more details). To obtain a more accurate estimate of action saliency, we instead measure the influence of an action on the observation processing result. This feature representation directly affects future action generation, as it serves as the input to the action expert. We introduce a lightweight predictor to estimate this feature space difference.

Specifically, the lightweight transformer-based predictor takes the image embedding (extracted by the ViT in the VLM), which represents the current environment, as input. The remaining actions are fed into the predictor as conditional control signals, following the conditioning scheme used in DiT~\cite{DiT}. The predictor is trained to estimate the change in image embeddings after executing the remaining actions, using a simple mean squared error (MSE) loss. During inference, a thresholding strategy is applied to avoid performing early observation for actions with high saliency.
We ensure that the additional overhead introduced by the predictor remains within an acceptable range. The runtime overhead, including the extra ViT processing, accounts for only about 5\% of the model inference time. Moreover, the training cost of the predictor is significantly lower than that of full fine-tuning of the entire model (e.g., VLASH). It is also worth noting that the lightweight predictor preserves flexibility and can readily augment existing techniques with adaptive capability.

\begin{table*}[th]
\centering
\renewcommand{\arraystretch}{1.25}
\caption{\textbf{Main results on the LIBERO benchmark.} We present the performance and efficiency of StreamingVLA and baseline methods. The $T_{\text{action}}$ and $T_{\text{halt}}$ are adopted to measure the speed and fluency of execution. The ``AFM'' and ``AEO'' represent the ``action flow matching'' and ``adaptive early observation'' technique. The ``NEO'' stands for the ``naive early observation''. The ``ANAO'' stands for the ``action norm based adaptive observation'' as mentioned in \cref{sec:method-early-observation}}
\label{tab:mainres}
\resizebox{0.98\linewidth}{!}{
\begin{tabular}{cccccccc}
\toprule[1pt]
\multirow{2}{*}{Method} & \multicolumn{5}{c}{Success Rate (\%) $\uparrow$} & Time per Action (ms) $\downarrow$ & Halting Gap (ms) $\downarrow$ \\
\cmidrule(lr){2-6}
& Spatial & Object & Goal & Long & Average & $T_{\text{action}}$ & $T_{\text{halt}}$ \\
\midrule \midrule
$\pi_{0.5}$ ($h=5$) & \textbf{98.8} & 98.2 & 98.0 & 92.4 & 96.9 & 74.5 & 232.3 \\
$\pi_{0.5}$ ($h=10$) & 97.4 & 98.2 & 96.2 & 88.6 & 95.1 & 49.9 ({\small 1.49$\times$}) & 230.8 ({\small 1.01$\times$}) \\
\hline
RTC (d=1)~\cite{RTC} & 96.2 & 19.8 & 93.0 & 25.2 & 58.55 & 50.2 ({\small 1.48$\times$}) & 203.6 ({\small 1.14$\times$})\\
SmolVLA~\cite{smolvla} & 97.0 & 99.0 & 97.0 & 90.0 & 95.8 & 51.1 ({\small 1.45$\times$}) & 180.7 ({\small 1.28$\times$}) \\
VLASH~\cite{vlash} & 97.5 & 99.2 & 97.3 & \textbf{94.6} & 97.1 & 40.6 ({\small 1.83$\times$}) & -  \\
Temporal Ensembling & 97.6 & 92.4 & 91.4 & 78.6 & 90.0 & 279.0 ({\small 0.26$\times$}) & 231.6 ({\small 1.003$\times$}) \\
\hline
\rowcolor{gray!15}
StreamingVLA \small{(AFM)} & 97.6 & \textbf{99.0} & \textbf{98.8} & 93.0 & \textbf{97.1} & 33.7 ({\small 2.21$\times$}) & 76.1 ({\small 3.05$\times$}) \\
\rowcolor{gray!15}
StreamingVLA \small{(AFM+NEO)} & 80.6 & 93.8 & 92.2 & 78.2 & 86.2 & \textbf{29.3 ({\small 2.54$\times$}) }& \textbf{23.0 ({\small 10.10$\times$})} \\
\rowcolor{gray!15}
StreamingVLA \small{(AFM+ANAO)}  & 86.0 & 96.2 & 93.0 & 84.8 & 90.0 & 30.775 ({\small 2.42$\times$}) & 27.75 ({\small 8.37$\times$}) \\
\rowcolor{gray!15}
StreamingVLA \small{(AFM+AEO)} & 96.6 & 96.6 & 95.4 & 91.0 & 94.9 & 31.625 ({\small 2.36$\times$}) & 36.0 ({\small 6.45$\times$}) \\


\bottomrule[1pt]
\end{tabular}}
\end{table*}

\section{Experiments}
\label{sec:exp}
\subsection{Implementation Details}

\noindent \textbf{Implementation Details: } We adopt \textbf{$\pi_{0.5}$-Libero}\cite{pi0.5, libero} as our base model and conducted training on four NVIDIA A800-SXM4-80GB GPUs. Models are trained for 60,000 iterations with a learning rate of 1e-5, taking 10-20 GPU hours to finish.  For the selection of the stabilizing factor $k$ and the variance $\sigma_0^2$ defined in \cref{sec:method-action-flow-matching}, we conduct grid search and select $k = 5$ and $\sigma_0^2 = 0.16$ for the subsequent training. Similarly, for the adaptive early observation predictor, we trained it for 54,000 iterations and employed a cosine learning rate schedule with an initial value of 1e-4. For the optimal threshold of the predictor, we evaluated both task success rate and inference speed across a range of values and ultimately selected the setting that achieves the best accuracy-latency trade-off. After training, models are evaluated on a single NVIDIA A800-SXM4-80GB GPU. All models are trained, deployed, and evaluated using the official PyTorch framework of OpenPI \cite{openpi2024}.

\noindent \textbf{Task Suites: } Following prior literaure~\cite{RTC, smolvla, pi0.5}, We select four task suites from the Libero simulation environment: \textit{Spatial}, \textit{Object}, \textit{Goal}, and \textit{Long} to evaluate model performance. Each suite consists of 10 distinct tasks.


\noindent \textbf{Baseline Methods: }
We adapt StreamingVLA to $\pi_{0.5}$-Libero, a state-of-the-art VLA model featuring a flow-matching action expert. By default, $\pi_{0.5}$-Libero uses $N_{\text{chunk}}=10$ and replans to 5 actions, meaning only the first five actions from each generated chunk are executed, corresponding to a horizon size of $h=5$. In action flow matching, actions are generated individually, so no replan is applied. For fair comparison, we also present results for $\pi_{0.5}$ without replan ($h=10$), obtained using the official codebase. We also apply four baseline methods on the $\pi_{0.5} (h=10)$ model, which also overlaps the latency of multiple stages: 
\begin{itemize}
    \item \textbf{Real-Time Chunking (RTC):}  
 measure inference delay $d = \frac{T_{o} + T_{g}}{T_{e}/h}$ during execution, and use it as the early observation steps $N_{eo}$. The next chunk begins with these $d$ actions already being executed, while the remaining $(h-d)$ actions are generated via an inpainting process, thereby enabling overlapped execution and generation. Following VLASH\cite{vlash}, we set a fix $d=1$ in our evaluation.
    \item \textbf{SmolVLA:}  
 perform an early observation before the current chunk finishes execution. The executed actions are removed from the new chunk.
    \item \textbf{VLASH:}
finetune the model with rectified states to mitigate misalignment caused by early observation.
    \item \textbf{Temporal Ensembling:}
maintain a buffer of predicted action chunks and executing, at each timestep, the average of all actions predicted for that timestep.

\end{itemize}
    
\noindent \textbf{Evaluation Metrics: } We perform 50 rollouts per task across all four suites, and evaluate the following metrics:

\begin{itemize}
    \item \textbf{Success Rate (\%):} The percentage of successful rollouts among all attempts.
    \item \textbf{Timing per action $T_{\text{action}}$ (ms):} The average time per executed action, computed as total time divided by horizon.
    \item \textbf{Halting time $T_{\text{halt}}$ (ms):} The average idle time between executions, measuring the fluency. 
\end{itemize}

\subsection{Experimental Results}

The comparison of efficiency-accuracy trade-off with baseline methods is presented in \cref{tab:mainres}, We conclude the key findings as follows: 

\noindent \textbf{(1) StreamingVLA effectively reduces latency \& halting Gap, achieving fast and fluent execution. } 
StreamingVLA achieves an average success rate of 94.9\%, compared with 95.1\% for $\pi_{0.5}$ (h=10), indicating negligible performance degradation of less than 0.5\%. In terms of overall latency, StreamingVLA yields a 1.57$\times$ speedup over $\pi_{0.5}$ (h=10) and 2.35$\times$ over $\pi_{0.5}$ (replan=5).  
Moreover, the Halting time is significantly reduced from around 232.3 ms to 36.0ms (a 6.45$\times$ improvement), demonstrating substantial improvement on execution fluency.
It is also worth noting that when adopting action flow matching (AFM) alone, the model improves success rates across all suites, while achieving 2.21$\times$ and 1.48$\times$ latency reductions and a 3.05$\times$ halting gap improvement compared to $\pi_{0.5}$ (h=5/10).

\noindent \textbf{(2) Baseline methods face challenges and limitations for speedup.} 

For the RTC baseline, the primary speed gain comes from a shorter halting gap (203.6 ms, 1.14×), which reduces the waiting time between observations and action execution. However, this early observation strategy leads to a pronounced drop in success rate, with the average performance falling to 58.55\% and long-horizon tasks degrading drastically. Although RTC attempts to refine the predicted velocity field through gradient-based correction during inference, the improvement is limited and insufficient to compensate for the information loss caused by premature observation. As seen in \cref{tab:mainres}, SmolVLA could moderately reduce halting from 230 to 180 ms, and the latency remain nearly unimproved (51.1 vs. 49.9). VLASH and our method optimize early observation from distinct dimensions. While training-based VLASH maintains success rates under acceleration, StreamingVLA achieves superior speed–performance trade-off by enabling parallel action generation and execution. The naive early observation (NEO), could significantly reduce latency and halting, but incur severe performance degradation (from 97.1 to 86.2), our adaptive early observation, could preserve the performance (94.7) while delivering substantial speedup.

\begin{table*}[ht]
    \centering
    \renewcommand{\arraystretch}{1.25}
    \caption{\textbf{Ablation studies for StreamingVLA techniques} validate the effectiveness of our Action Flow Matching(AFM) formulation w/o state-based alignment modeling(SA), normalization modification(NM) mentioned in \cref{sec:method-action-flow-matching}, as well as adaptive early observation(EO) schemes. The ``2'' and ``1.45'' represents the average number of early observed actions. The "Random" represents employing early observation randomly.}
    \label{tab:ablation}
    \resizebox{0.97\linewidth}{!}{
        \begin{tabular}{cccccccc}
        \toprule[1pt]
        AFM & NM & SA & Horizon & EO & Success Rate(\%) & $T_{\text{action}}$ (ms) &  $T_{\text{halt}}$ (ms)\\
        \midrule\midrule
        - & - & - & 10 & - & 95.1 & 49.9 & 230.78\\
        \checkmark & - & - & 10 & - & - & - & -\\
        \checkmark & \checkmark & - & 10 & - & 61.8 & - & - \\
        \checkmark & \checkmark & \checkmark & 10 & - & 97.1 & 33.7 & 76.1\\
        \checkmark & \checkmark & \checkmark & 10 & Naive (2) & 86.2 & 29.3 & 23.0\\
        \checkmark & \checkmark & \checkmark & 10 & Random (1.4) & 90.875 & 32.125 & 38.3 \\ 
        \rowcolor{gray!15}
        \checkmark & \checkmark & \checkmark & 10 & Adaptive (1.4) & 94.9 & 31.625 & 36.0\\
        \bottomrule[1pt]
        \end{tabular}}
\end{table*}

\section{Ablation Studies}
\label{sec:ablation}
To validate the effectiveness of the proposed techniques, we present the results in \cref{tab:ablation}. The key findings are summarized as follows: 

\noindent \textbf{(1) Effectiveness of Action Flow Matching (AFM).} Replacing the vanilla formulation with AFM significantly improves temporal modeling of actions. Although AFM alone does not directly report full metrics in the table, its combination with subsequent components consistently yields better stability and performance, indicating that modeling action evolution as a continuous flow provides a more structured optimization objective.

\noindent \textbf{(2) Critical role of Normalization Modification (NM).} Normalization Modification is a key component of the new AFM formulation. Without NM, the model fails to train effectively under the action flow parameterization. As shown in the table, introducing AFM without proper normalization does not yield stable or competitive results, while enabling NM establishes a well-conditioned training objective. Although NM alone (without SA) leads to poor final performance (61.8\% success rate), this does not indicate ineffectiveness; rather, it shows that NM is a necessary but not sufficient component. In other words, NM provides the essential scaling and stabilization required for optimizing the continuous action flow, and meaningful performance can only be achieved when NM is combined with state-based alignment.

\noindent \textbf{(3)Importance of State-based Alignment (SA).} Adding SA on top of AFM and NM dramatically improves the success rate to 97.1\%, surpassing the baseline (95.1\%), while simultaneously reducing both action inference time ($T_{\text{action}}$: 49.9 → 33.7 ms) and halting time ($T_{\text{halt}}$: 230.78 → 76.1 ms). This verifies that maintaining consistency between action-space state and physical state is crucial for both accuracy and efficiency.

\noindent \textbf{(4)Early Observation (EO) for latency reduction.} Naive EO aggressively exposes two future actions at each step, achieving the lowest latency but causing a clear performance drop (86.2\%), indicating that excessive early observation harms decision quality.

Random and Adaptive EO share the same average number of early observed actions (1.4), ensuring a fair comparison. While Random EO provides a reasonable speed–performance trade-off, Adaptive EO further improves success rate (94.9\% vs. 90.9\%) by leveraging an adaptor to guide when early observation should be triggered, demonstrating that informed scheduling is more effective than unguided randomness.

Overall, the ablation study confirms that (1) AFM provides a principled action modeling framework, (2) SA is essential for stabilizing AFM-based training, and (3) Adaptive Early Observation effectively reduces latency while largely preserving task success, validating the design of StreamingVLA.

\section{Conclusions}
\label{sec:conclusion}

To address the stuttered and haltinbg execution issue of VLA, we present StreamingVLA, which enables parallel processing of multiple stages in VLA in a streaming manner, without sacrificing performance. It achieves a 2.4× speedup in latency and a 6.5× reduction in halting time, enabling fast and fluent execution.

%
%
\bibliographystyle{splncs04}
\bibliography{main}
\clearpage
\appendix
\section*{Appendix}
\section{Real-World Robot Deployment}
\label{sec:deploy}

\begin{figure*}[h]
    \centering
    \includegraphics[width=0.77\linewidth]{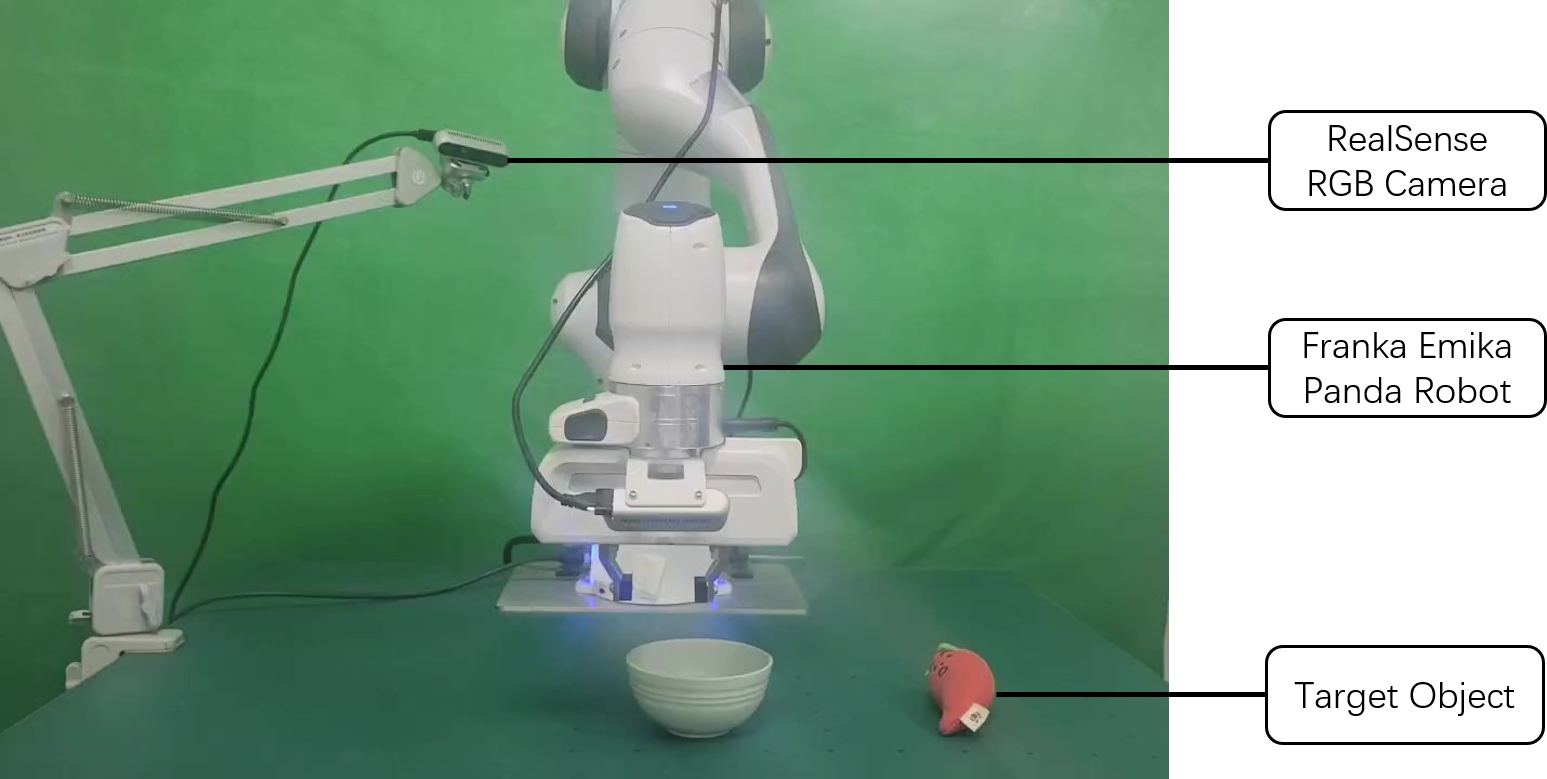}
    \caption{ \textbf{Real-world Setup.} The real-world evaluation platform includes a tabletop workspace, a Franka Panda robotic manipulator fixed to the table, and a RGB camera for visual observation. All objects are positioned on the table surface.}
    \label{fig:setup}
\end{figure*}

To validate the practical efficiency of our method, we deploy the model on a real-world robotic manipulation task. The task requires a robotic arm to pick up an object from a designated location and place it at a target position. This pick-and-place scenario represents a common manipulation setup that requires precise action execution and timely perception updates.

Our model is trained based on the \textbf{$\pi_{0.5}-$base} architecture with an action horizon of $H=8$. During deployment, we compare our \textbf{StreamingVLA} framework with the original $\pi_{0.5}$ policy under the same hardware setup. For the baseline configuration, the $\pi_{0.5}$ policy operates with an action horizon of $H=8$ and a replanning step of 4 actions, while \textbf{StreamingVLA} uses the same horizon but performs streaming action generation.

Both methods are deployed on the same robotic platform as \cref{fig:setup} to ensure a fair comparison. The system consists of a table-top workspace with a Franka Emika Panda robot and a fixed RGB camera, whose captured images serve as visual input to the VLA model. The robot has seven actuated joints and a parallel-jaw gripper with open/close capability. We measure the average time per action demonstrated in \cref{sec:exp} during the manipulation process. The results show that \textbf{StreamingVLA} achieves an average latency of \textbf{170.88 ms} per action, while the original $\pi_{0.5}$ baseline requires \textbf{271.49 ms} per action.

These results demonstrate that \textbf{StreamingVLA} significantly improves real-world control efficiency by reducing action generation latency while maintaining the same action horizon. The reduced latency enables faster response during execution, which is critical for real-time robotic manipulation. 

\section{Algorithms}
\label{sec:alg}

We introduce the action-space state $\alpha$ as a new type of input during training. These states are directly read from the preprocessed dataset and used as the initial action state $\alpha_i$of each sub-trajectory $\xi_i$, from which the ground-truth action states of the entire sub-trajectory $\mathcal{A}_i$ are computed. Consequently, our training algorithm differs from that of the streaming flow policy or $\pi_{0.5}$; We present the detailed description in \cref{alg:training} as follows:

\begin{algorithm}[t]
    \caption{Training algorithm}
    \begin{algorithmic}[1]
            {\Require Preprocessed training set $\mathcal{D} = \{(o_i,\alpha_i,\xi_i)\}_{i=1}^N$}, horizon $h$, noise scale $\sigma_0$, stabilizing factor $k$
            \While{\text{not converged}}
                \State $(o_i,\alpha_i,\xi_i) \sim \mathcal{D}$
                \State Compute action state $\mathcal{A}_i:$
                \State $\mathcal{A}_i[n] \leftarrow \mathcal{A}_i[n-1]+\xi_i[n-1],\ \mathcal{A}_i[0] \leftarrow \alpha_i$
                \State $t \sim \mathrm{Uniform}(0, 1)$, index $T \leftarrow \lfloor t\cdot h\rfloor$
                \State $x_T \sim \mathcal{N}(\mathcal{A}_i[T],\sigma_0^2e^{-2kT/h})$
                \State $\theta \leftarrow \theta - \hspace{0em} \lambda \nabla_\theta \hspace{-0.2em} \underbrace{\|v_\xi(x_T, T/h) - v_\theta(x_T, T/h \vert o_i)\|^2}_{{\text{Conditional flow matching loss}}}$
            \EndWhile
            \State \Return $v_\theta$
        \end{algorithmic}
        \label{alg:training}
\end{algorithm}

During inference we also need to maintain the action state, which is updated in real time according to the actions that have been executed. The corresponding procedure is described in \cref{alg:inference}.
\begin{algorithm}[t]
            \caption{\small Inference algorithm}
            \begin{algorithmic}[1]
                {\Require $v_\theta(x_T, T/h \vert o)$}, Early observation step $N_{eo}$, Observation indicator $I$ with threshold $\eta$
                \State $o\leftarrow o_0,\ \alpha \leftarrow \textbf{0}, \ T\leftarrow 0,\ \text{need\_obs} \leftarrow \text{True}$ \quad//initialization
                \While{True}
                    \If{need\_obs}
                        \State $o\leftarrow$ Observation \quad//async process observation
                        \State need\_obs $\leftarrow$ False
                    \EndIf
                    \State action $a \leftarrow v_\theta(\alpha,T/h\vert o)\frac{1}{h}$ \quad//async generate action
                    \If{$h-T=N_{oe}$}
                        \If{$I(a)\leq \eta$}\quad//adaptive early observation
                            \State need\_obs $\leftarrow$ True 
                        \EndIf
                    \EndIf
                    \If{$h-T=0$}
                        \State need\_obs $\leftarrow$ True
                        \State $T\leftarrow0$
                    \EndIf
                    \State execute($a$)\quad//async execute action
                    \State $\alpha \leftarrow \alpha + a$\quad//update action state
                    \State $T \leftarrow T+1$
                \EndWhile
            \end{algorithmic}
            \label{alg:inference}
        \end{algorithm}
        
\section{Additional Experimental Results}
\label{sec:add-res}

\subsection{Runtime Breakdown}
\label{sec:runtime}

\begin{figure*}[t]
    \centering
    \includegraphics[width=0.77\linewidth]{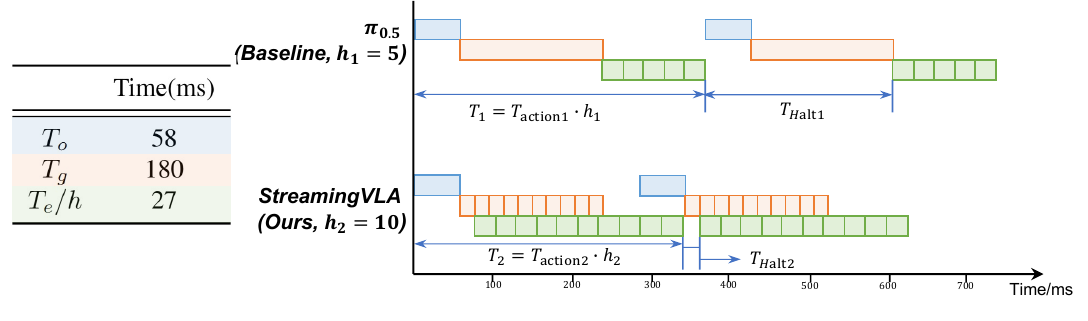}
    \caption{\textbf{The runtime breakdown based on actual profiling of $\pi_{0.5}$ and StreamingVLA.} The three colored blocks correspond to the “observation”, "gen-action" and "execution" stage in the VLA workflow. The length of each block reflects the measured time.}
    \label{fig:timeline_overall}
\end{figure*}

We further analyze the runtime composition based on actual profiling using the analysis framework discussed in \cref{sec:method-runtime-analysis} and illustrated in \cref{fig:timeline_overall}, to evaluate the baseline method and identify the sources of the speedup in StreamingVLA.

For \textbf{$\pi_{0.5}$}, no acceleration methods are applied thus $O_{ge}=O_{oe}=0$, the latency and halting time could be calculated as follows, matching the measured latencies: 
\begin{align*}
    T_{\text{action1}} &= \frac{T_o + T_g + h_1 T_e}{h_1} = 74.6 \text{ms}, \\
    T_{\text{halt1}} &= T_o + T_g = 238 \text{ms}.
\end{align*}
For \textbf{StreamingVLA}, The overlap time $O_{ge}$ and $O_{oe}$ can be calculated as below:
\begin{align*}
    O_{ge} &= (h-1)\min\left\{\frac{T_g}{h},\frac{T_e}{h}\right\} = 162 \text{ms}, \\
    O_{oe} &= N_{eo}\frac{T_e}{h} = 41.58 \text{ms}.
\end{align*}
Then we obtain:
\begin{align*}
    T_{\text{action2}} & = \frac{T_o + T_g + T_e - O_{ge}-O_{oe}}{h} = 30.44 \text{ms}, \\
    T_{\text{halt2}} & = T_o + T_g - (O_{ge}+O_{oe}) = 34.42 \text{ms}.
\end{align*}
In summary, StreamingVLA delivers a 2.40$\times$ latency and 6.71$\times$ halting gap improvements.

\subsection{Design Details of Early Observation Predictor}
\label{predictor}

\begin{figure*}[t]
    \centering
    \includegraphics[width=0.77\linewidth]{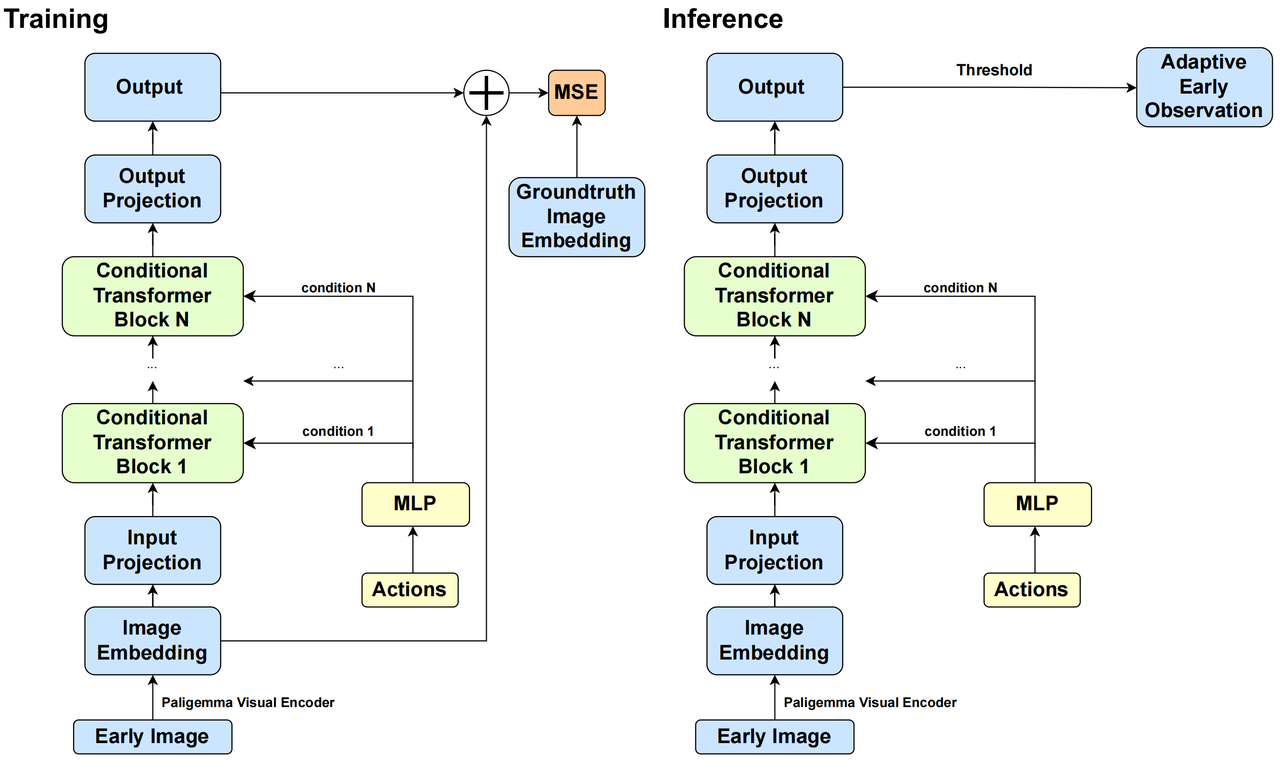}
    \caption{\textbf{Architecture of the predictor for adaptive early observation.} During training, the model receives the early-frame embedding and the pending action sequence as conditions and predicts a residual embedding $\Delta embedding$ that estimates the latent change toward the future frame. The model is trained using an MSE loss between the ground-truth late-frame embedding and the reconstructed representation ($embedding_{early} + \Delta embedding$). At inference time, the $L_2$ norm of $\Delta embedding$ serves as a visual-variance indicator, which is compared with a threshold to determine whether early observation should be triggered.}
    \label{fig:predictor}
\end{figure*}

The architecture of our predictor model is illustrated in \cref{fig:predictor}. We utilized two NVIDIA A800-SXM4-80GB GPUs to train the predictor, taking 3-4 GPU hours to finish. During training, we randomly sample frame pairs with a temporal gap of 1–3 actions. A visual encoder first maps these observations into an embedding space. The predictor then takes the early frame embedding as input and the pending action sequence as condition to generate a residual embedding, $\Delta embedding$. The model is optimized using a Mean Squared Error (MSE) loss between the latent representation of the late frame and the predicted composite ($embedding_{early} + \Delta embedding$). 

During inference, the $L_2$-norm of the predicted $\Delta embedding$ serves as the metric for visual variance. Early observation is inhibited if this value exceeds a predefined threshold, and we select this threshold by analyzing the accuracy-latency tradeoff across various settings. 

In our framework, each observation generates a horizon of $H$ actions ($H=10$ in our setting), which are appended to the pending execution queue. When the queue reduces to $n$ remaining actions (a predefined parameter), the current observation and residual actions are transmitted to the server. The predictor in the server then evaluates whether to trigger an early observation. 

Consequently, the predictor operates once per $H$ actions, introducing a marginal latency of 8–10 ms per invocation. This translates to an average computational overhead of approximately 0.8–1 ms per action, representing only 5\% of the standard generation time (18 ms). Despite this minimal overhead, our method achieves a significant accuracy improvement of 8.7\% compared to the naive early observation baseline.

\subsection{Visualization}
\label{sec:visualization}
We visualized representative tasks from each suite to provide an intuitive comparison of task completion speed and fluency between our method and the baselines. For each suite, we selected one representative task for visualization. The tasks corresponding to the four suites are as follows:

\begin{itemize}
    \item \textbf{LIBERO-Spatial:} pick up the black bowl in the top drawer of the wooden cabinet and place it on the plate.
    \item \textbf{LIBERO-Object:}pick up the orange juice and place it in the basket.
    \item \textbf{LIBERO-Goal:} open the top drawer and put the bowl inside.
    \item \textbf{LIBERO-Long:} put both the alphabet soup and the tomato sauce in the basket.
\end{itemize}

\begin{figure*}[t]
    \centering
    \includegraphics[width=1.0\linewidth]{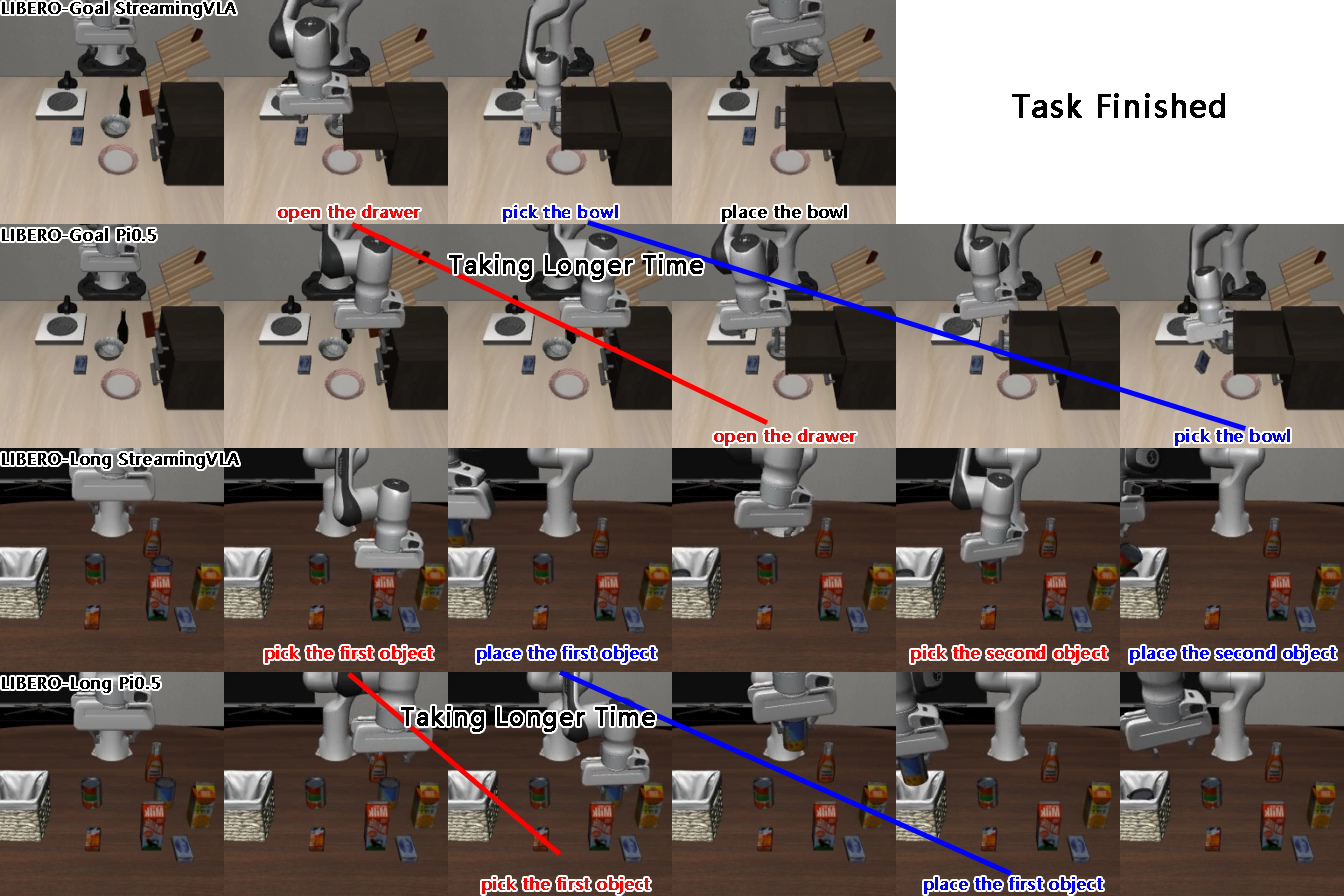}
    \caption{\textbf{Screenshots of our video simulation results.}The images in each row are spaced 4 seconds apart. \textbf{Top two rows:} \textit{open the top drawer and put the bowl inside.} StreamingVLA completes the task in under 16 seconds, whereas pi0.5 has not yet picked up the bowl at the 20-second mark. \textbf{Bottom two rows:} \textit{put both the alphabet soup and the tomato sauce in the basket.} At 20 seconds, StreamingVLA has already finished the pick-and-place operations for both objects, while pi0.5 is still on its way to picking up the second item.}
    \label{fig:sim}
\end{figure*}

We recorded the simulation state every 30 ms (approximately the time required to execute one action step in the LIBERO environment), and used these snapshots to generate video results. A subset of these results is shown in \cref{fig:sim}. During the “open the top drawer and put the bowl inside” task, the robotic arm controlled by the $\pi_{0.5}$ model experienced 12 pauses while opening the drawer, and even failed to open it once due to a prolonged pause. When picking up the bowl and placing it inside the drawer, the arm encountered an additional 25 pauses. These interruptions significantly delayed the task completion and even hindered its success. In contrast, the robotic arm controlled by StreamingVLA exhibited no noticeable halting throughout the execution of this task. 

\end{document}